\documentclass[11pt,a4paper]{article}
\usepackage{times,latexsym}
\usepackage{url}
\usepackage[utf8]{inputenc}
\usepackage[T1]{fontenc}
\usepackage[acceptedWithA]{tacl2018v2}
\usepackage{fancyhdr}
\fancypagestyle{taclpub}{%
  \fancyhf{}%
  \fancyfoot[C]{\footnotesize Published in \emph{Transactions of the Association
  for Computational Linguistics}, 14:1787--1802, 2026.
  \url{https://doi.org/10.1162/TACL.a.774}. Author's final version.
  \copyright~2026 Association for Computational Linguistics, CC BY 4.0.}%
}
\usepackage{graphicx}
\usepackage{amsmath,amssymb,amsfonts}
\usepackage{booktabs}
\usepackage{enumitem}
\usepackage{placeins}
\usepackage{balance}


\begin{document}

\title{SupraTok: Cross-Boundary Tokenization for Enhanced Language Model Performance}

\author{
  Andrei-Valentin Tănase \and Elena Pelican
  \\
  Faculty of Mathematics and Computer Science
  \\
  ``Ovidius'' University of Constanta, Romania
  \\
  \texttt{valentin.tanase@365.univ-ovidius.ro}
  \\
  \texttt{epelican@365.univ-ovidius.ro}
}

\date{}

\maketitle

\thispagestyle{taclpub}

\begin{abstract}
  Tokenization remains a persistent bottleneck in language modeling, especially when vocabulary learning is limited by whitespace boundaries. We present SupraTok, a tokenizer that crosses whitespace boundaries using three modular components: optional entropy-based data curation, staged curriculum training with PMI-guided candidate search, and multilingual script handling. At 100k vocabulary on the same unfiltered training data, SupraTok improves compression over standard BPE by 17.5\% and over the official SuperBPE implementation by 1.8\%, while training 2.1$\times$ faster than SuperBPE. Across 50k--300k vocabularies in the same matched setting, SupraTok remains ahead of SuperBPE by 1.8\%--8.6\%. We evaluate entropy filtering separately as a pipeline step: at 100k vocabulary it raises SupraTok from 5.78 to 5.99 C/T, while matched controls show a smaller gain for SuperBPE and almost no change for SP-BPE-CrossBoundary. On FLORES-200 across 14 languages, SupraTok yields a macro-averaged 34.9\% relative gain over the BPE baseline. In separate downstream experiments with matched compute and fixed token budgets, using 12L-768d and 24L-1024d GPT-2-style backbones with 256k vocabularies, SupraTok improves HellaSwag and MMLU. Overall, these results show that crossing whitespace boundaries gives consistent compression gains under controlled public-data comparisons, while optional entropy filtering provides a separate pipeline benefit.
\end{abstract}

\section{Introduction}

Recent progress in language modeling has been driven primarily by increases in model scale, training data, and architectural refinement~\citep{ref6,ref7,ref8}. By comparison, tokenization remains relatively underexplored, even though it directly controls sequence length, vocabulary allocation, and the interface between raw text and model computation.

Tokenization serves as the interface between text and neural computation. Modern language models commonly use subword tokenization methods such as BPE, WordPiece, and unigram language-model tokenizers, often implemented through toolkits such as SentencePiece~\citep{ref9,ref10,ref11,ref12,ref13}. In practice, these tokenizers are frequently trained with whitespace-aware pretokenization, which prevents tokens from crossing word boundaries.

Whitespace-aware tokenization is convenient for many alphabetic languages, but it can be restrictive when useful units cross whitespace boundaries, when scripts do not mark boundaries consistently, or when recurring multiword expressions behave as reusable lexical units~\citep{ref14,ref15}. These restrictions increase sequence length and consume vocabulary capacity on fragments that may be less useful than reusable cross-boundary units.

Recent research has explored alternatives. Byte-level and character-level approaches relax or eliminate tokenization constraints, but usually at the cost of architectural changes or much longer sequences~\citep{ref17,ref18,ref19}. Concurrent work on SuperBPE~\citep{ref20} and Boundless BPE~\citep{ref26} shows that cross-boundary tokenization can improve compression substantially. Our contribution is a modular tokenizer architecture with three components: optional entropy-based data curation, a staged training curriculum with PMI-guided candidate discovery, and script-aware processing for multilingual settings, evaluated under matched-data comparisons.

We therefore focus on two questions. First, how much of SupraTok's compression gain comes from the tokenizer algorithm itself under controlled unfiltered training data? Second, how much additional benefit comes from optional entropy-based curation as a separate pipeline component? The remainder of the paper is organized to answer those questions directly while also evaluating multilingual compression and downstream language-model performance.

\section{Related Work}

\subsection{Subword Tokenization}

Modern language models commonly use subword tokenization methods such as BPE, WordPiece, and unigram language-model tokenizers~\citep{ref9,ref10,ref11,ref13}. SentencePiece~\citep{ref12} provides a widely used toolkit for both BPE and unigram training directly from raw text, including optional control over whitespace splitting.

\subsection{Cross-Boundary and N-gram Tokenization}

Recent work has begun to relax the standard whitespace-boundary assumption in different ways. SuperBPE~\citep{ref20} introduces a two-stage pretokenization curriculum that first learns conventional subwords and then resumes training while lifting whitespace restrictions to learn superwords. SentencePiece~\citep{ref12} supports cross-boundary training through \texttt{split\_by\_whitespace=false}, which we include as a direct baseline. Boundless BPE~\citep{ref26} is also relevant as concurrent work on tokenizer construction without pretokenization boundaries. Our contribution is therefore not the first cross-boundary tokenizer, but a modular tokenizer architecture that combines controlled curriculum learning, PMI-guided candidate discovery, and optional entropy-based data curation under a fair-comparison protocol.

\textbf{Relationship to SuperBPE.} SuperBPE and SupraTok are best viewed as two different curriculum-based approaches to cross-boundary tokenization. In this paper, we compare them under matched public-data conditions rather than contrasting a curriculum method with a non-curriculum one. Under identical unfiltered training data, SupraTok improves compression over SuperBPE at 100k vocabulary and remains ahead across a 50k--300k vocabulary sweep. We additionally evaluate the same entropy-filtering procedure separately on the cross-boundary tokenizers in Table 4B to distinguish algorithmic gains from pipeline-level preprocessing gains.

\subsection{Tokenization Scaling Laws}

Recent work has investigated the relationship between vocabulary size and model performance. \citet{ref27} demonstrate that optimal vocabulary size scales with model size and training data, finding that larger models benefit from larger vocabularies. Their analysis shows diminishing returns beyond certain vocabulary sizes relative to model capacity, suggesting careful vocabulary optimization is essential.

\citet{ref28} extend this analysis to n-gram tokenization, showing that incorporating multi-token units (analogous to our cross-boundary patterns) can improve compression efficiency and downstream performance. Their scaling analysis reveals that n-gram vocabulary benefits compound with model size, supporting our hypothesis that semantic unit preservation becomes increasingly valuable at scale.

SupraTok contributes to this line of research by showing that which units occupy the vocabulary, especially cross-boundary semantic units rather than arbitrary fragments, can materially affect compression alongside vocabulary size itself. Our PMI-guided pattern selection provides a principled approach to vocabulary composition that complements scaling law insights.

\subsection{Linguistically-Motivated Tokenization}

Prior work has incorporated linguistic knowledge into tokenization through methods such as Morfessor-style morphological analysis~\citep{ref29}, stochastic regularization via BPE-dropout~\citep{ref30}, and multilingual morphological features~\citep{ref31}. SupraTok differs by focusing on cross-word rather than within-word units---identifying multi-word expressions that linguistic theory recognizes as fundamental to language processing~\citep{ref21}.

\section{Methods}

\subsection{Overview of the SupraTok Pipeline}

SupraTok consists of three modular components. First, optional entropy-based data curation filters low-information documents from the tokenizer-training corpus. Second, the tokenizer is trained through three sequential passes that progressively relax boundary constraints and expose richer cross-boundary candidates. Third, SupraTok uses script-aware processing, with additional language-specific heuristics for multilingual settings. We describe these components separately and evaluate them separately in Section~4.

Throughout the paper, we measure compression as characters per token (C/T), where larger values indicate better compression.

\subsection{Entropy-Based Data Curation}

The quality of training data affects which patterns dominate tokenizer learning. SupraTok therefore supports an optional entropy-based filtering stage before vocabulary training.

We calculate character-bigram entropy for each document in the training corpus:

\begin{equation}
H(\text{doc}) = -\sum_i p(b_i) \log_2 p(b_i)
\end{equation}

where $b_i$ represents character bigrams. Documents are classified into three categories based on their entropy: low ($H < 3.0$), medium ($3.0 \leq H \leq 4.5$), and high ($H > 4.5$). Low-entropy documents tend to contain boilerplate, repetitive content, or data artifacts; high-entropy documents are typically more information-dense.

We retain 10\% of low-entropy, 50\% of medium-entropy, and 90\% of high-entropy documents. In the 100k-document English setup used for the matched filtering comparison, this procedure yields a 76k-document filtered subset. The filtering stage is optional and is evaluated separately from the algorithm-only comparison.

\subsection{Cross-Boundary Vocabulary Learning}

SupraTok uses three ordered passes but no fixed per-pass vocabulary quotas. For a target vocabulary size $V$, training begins with \textsc{traditional} merges, then exposes PMI-seeded \textsc{multiword} candidates, and finally considers longer \textsc{expression} spans. Each pass contributes only the additions that survive its own candidate-generation and acceptance rules, and later passes continue from the current vocabulary rather than restarting from scratch. Training stops when $|\mathcal{T}| = V$ or no valid candidates remain.

Candidate multiword patterns are seeded using pointwise mutual information (PMI):

\begin{equation}
\text{PMI}(w_1, \ldots, w_n) = \log_2\left(\frac{P(w_1,\ldots,w_n)}{\prod_i P(w_i)}\right)
\end{equation}

Seed mining considers $n \in \{2,\dots,6\}$, with PMI threshold 2.0 and minimum frequency 100. We also apply overlap controls so that the initial seed set favors broad coverage rather than many near-duplicate variants of the same phrase.

In Phase 3, SupraTok evaluates candidate spans of length 2--5 using a composite score that combines held-out token-reduction potential, PMI/NPMI-based cohesion, boundary entropy, lexical specificity, and penalties for stopword-heavy or malformed spans. Final acceptance is determined by a post-hoc leaf-surgery rule: a span is added only when its estimated held-out utility exceeds the utility of the merges it would displace. Appendix~\ref{app:algo} gives the English Phase 3 scoring form, thresholds, and rejection rules.

To model boundary consistency, we use left and right branching entropy; the left-entropy term is

\begin{equation}
H_{\text{left}}(w) = -\sum_c P(c|w) \log_2 P(c|w)
\end{equation}

where $c$ represents possible left contexts for sequence $w$. Low branching entropy indicates that a sequence appears in stable contexts and is therefore a stronger candidate lexical unit.

\subsubsection{Algorithm Overview}

Figure~\ref{alg:supratok} presents the complete SupraTok training procedure.

\begin{figure}[t]
\scriptsize
\begin{tabular}{p{7.0cm}}
\hline
\textbf{SupraTok Training} \\
\hline
\textbf{Input:} Corpus $\mathcal{D}$, target vocabulary $V$, PMI threshold $\tau$ \\
\textbf{Output:} Tokenizer vocabulary $\mathcal{T}$ \\
\hline
1: Optionally filter $\mathcal{D}$ with entropy-based data curation to obtain $\mathcal{D}'$ \\
2: Mine seed phrases on $\mathcal{D}'$ using n-grams with $n \in \{2,\dots,6\}$ and PMI threshold $\tau$ \\
3: Initialize $\mathcal{T}$ with the character alphabet, byte-fallback symbols, and selected seeds \\
4: \textbf{for} pass $\in \{\textsc{traditional}, \textsc{multiword}, \textsc{expression}\}$ \textbf{do} \\
5: \quad Construct the pass-specific candidate set $Q_{\mathrm{pass}}$ under the current boundary rules \\
6: \quad \textbf{while} $|\mathcal{T}| < V$ and $Q_{\mathrm{pass}}$ is not empty \textbf{do} \\
7: \qquad \textbf{if} pass $< \textsc{expression}$, select the best valid merge from $Q_{\mathrm{pass}}$ and apply it \\
8: \qquad \textbf{else} score candidate spans and accept only spans with positive net utility under leaf surgery \\
9: \textbf{return} $\mathcal{T}$ \\
\hline
\end{tabular}
\caption{SupraTok training procedure.}
\label{alg:supratok}
\end{figure}

Each pass is defined by its candidate visibility and acceptance rules, not by a preset vocabulary allotment.

\FloatBarrier
\subsection{Script-Aware Processing}

SupraTok applies Unicode-aware processing that adapts to broad script and language-type categories, with additional language-specific heuristics for multilingual settings. We classify Unicode characters into five supercategories: Letters and Marks (LM), Punctuation and Symbols (PS), Numbers (N), Separators (Z), and Control and Other (C). For scripts without explicit word boundaries, we gate merges using thresholded character-bigram evidence before allowing cross-boundary merges. For Turkish and Finnish, we approximate likely morphological boundaries by suffix matching against heuristic suffix lists mined from the multilingual tokenizer-training subset; this is a lightweight heuristic rather than a full morphological analyzer.

\subsubsection{Script-Aware Processing Validation}

To quantify the contribution of script-aware processing, we trained tokenizer variants with and without language-specific adaptations on identical multilingual data. Table~\ref{tab:script-aware} reports a focused seven-language ablation chosen to stress boundary-free scripts and morphology (Chinese, Japanese, Thai, Korean, Turkish, Finnish, and Arabic), using a 100k-vocabulary tokenizer trained on balanced CulturaX data (8,000 documents per language). The `Without' column disables script-specific adaptations while retaining all other SupraTok components.

Script-aware processing provides a macro-averaged 11.9\% relative compression improvement. The strongest script-aware gains appear for Japanese and Thai, where character-level gating helps identify reusable character sequences, and for Turkish and Finnish, where suffix-aware weighting improves morpheme-sensitive segmentation; Chinese, Korean, and Arabic also show consistent improvements.

\begin{table}[t]
\centering
\footnotesize
\begin{tabular*}{\columnwidth}{@{\extracolsep{\fill}}lccc@{}}
\toprule
\textbf{Language} & \textbf{Without} & \textbf{With} & \textbf{Relative difference (\%)} \\
\midrule
Chinese & 1.21 & 1.36 & 12.4\% \\
Japanese & 1.58 & 1.83 & 15.8\% \\
Thai & 2.34 & 2.71 & 15.8\% \\
Korean & 2.09 & 2.26 & 8.1\% \\
Turkish & 4.89 & 5.42 & 10.8\% \\
Finnish & 4.71 & 5.18 & 10.0\% \\
Arabic & 3.12 & 3.45 & 10.6\% \\
\midrule
\textbf{Average} & 2.85 & 3.17 & 11.9\% \\
\bottomrule
\end{tabular*}
\caption{Script-aware processing effect on compression (C/T) across seven languages; the average percentage is the macro-average of language-wise relative gains.}
\label{tab:script-aware}
\end{table}

Table~\ref{tab:script-aware} is a within-SupraTok ablation rather than a cross-method comparison: all rows compare SupraTok with versus without script-aware adaptations on identical multilingual data, while direct comparisons against BPE-family baselines are deferred to Table~\ref{tab:multilingual}. The average percentage is the macro-average of per-language relative gains. For Chinese, Japanese, and Thai, the relevant mechanism is thresholded character-bigram gating with a minimum count of 50; for Turkish and Finnish, it is suffix-aware weighting based on 200 heuristic suffix candidates per language mined from the multilingual training subset; Arabic applies hamza normalization before frequency counting. The separate 14-language benchmark in Table~\ref{tab:multilingual} uses balanced CulturaX subsets of 8,000 documents per language, sampled heuristically without additional stratification to keep the benchmark budget manageable while preserving cross-language balance; Appendix Table E1 provides the corresponding document, sentence, and Phase-1 token counts.

\FloatBarrier
\subsection{Hyperparameter Selection}

For English, Optuna searches three analysis-specific hyperparameters: target vocabulary size, later-pass candidate-screening PMI threshold, and later-pass candidate-screening frequency threshold. We rank each trial with a scalar score that rewards held-out compression and penalizes validation-LM perplexity and encoding latency. Compression is measured as characters per token on held-out text, perplexity is measured by a lightweight validation LM trained on the same English development setup, and latency is measured by tokenizer encoding time on a fixed held-out shard. Appendix~\ref{app:optuna} gives the search ranges and reference definitions used in the ranking score.

For a trial $t$ with target vocabulary size $V$, let $C_t$, $P_t$, and $L_t$ denote its compression, perplexity, and latency, and let $C_{\mathrm{ref}}(V)$, $P_{\mathrm{ref}}(V)$, and $L_{\mathrm{ref}}(V)$ denote the corresponding values of the BPE-Standard baseline trained on the same training subset and evaluated under the same protocol at vocabulary size $V$. We define
\begin{equation*}
\begin{aligned}
C_{\mathrm{norm}}(t) &= \frac{C_t-C_{\mathrm{ref}}(V)}{|C_{\mathrm{ref}}(V)|}, \\
P_{\mathrm{norm}}(t) &= \frac{\log_2(P_t)-\log_2(P_{\mathrm{ref}}(V))}{|\log_2(P_{\mathrm{ref}}(V))|}, \\
L_{\mathrm{norm}}(t) &= \frac{\log_2(L_t)-\log_2(L_{\mathrm{ref}}(V))}{|\log_2(L_{\mathrm{ref}}(V))|}.
\end{aligned}
\end{equation*}
\begin{equation}
\mathrm{Score}(t)=\alpha\,C_{\mathrm{norm}}(t)-\beta\,P_{\mathrm{norm}}(t)-\gamma\,L_{\mathrm{norm}}(t).
\end{equation}
We set $\alpha=\beta=\gamma=1$. All logarithms are base 2. Across 100 English candidate configurations, the highest-scoring configuration used a 256k target vocabulary, which we reuse in the downstream experiments. In the production English benchmark and ablation runs reported here, later-pass merge screening uses a candidate-screening PMI threshold of 5.0 and a candidate-screening frequency threshold of 5. Outside that vocabulary-selection step, the core SupraTok tokenizer settings are held fixed across experiments; the main analysis-specific differences are the training corpus, corpus size, and whether the tokenizer-training corpus is left unfiltered or first filtered by the entropy-based document-selection step before tokenizer training. The multilingual experiments likewise reuse the same core settings, with additional script-aware processing as described in Section~3.4.

\subsection{Implementation Notes}

SupraTok is implemented as a modular Python library built on HuggingFace Tokenizers. Key components include the main SupraTok trainer coordinating the three-pass curriculum, a cross-boundary pretokenizer for candidate discovery, a streaming data loader for entropy-filtered corpora, and an n-gram miner for PMI-based seed extraction. The resulting tokenizer can be exported to standard HuggingFace and SentencePiece-compatible formats.

Table~\ref{tab:compute} summarizes the computational requirements for downstream language-model training. Exact parameter totals for the 256k-vocabulary runs are reported in Appendix~\ref{app:param}.

\begin{table}[t]
\centering
\footnotesize
\begin{tabular*}{\columnwidth}{@{\extracolsep{\fill}}lccc@{}}
\toprule
\textbf{Backbone} & \textbf{Time} & \textbf{Hardware} & \textbf{CO$_2$} \\
\midrule
12L-768d & 21h & 8$\times$A100 & 24 kg \\
24L-1024d & 79h & 8$\times$A100 & 90 kg \\
\bottomrule
\end{tabular*}
\caption{Language model training requirements for downstream evaluation.}
\label{tab:compute}
\end{table}

\FloatBarrier
\subsection{Evaluation Protocol}

Compression is evaluated on WikiText-103 validation for English experiments and FLORES-200 devtest for multilingual experiments. Downstream language models are trained with nanoGPT on FineWeb-Edu using identical optimizer and backbone settings across tokenizer conditions. Section~\ref{sec:exp-method} gives the full fair-comparison protocol, baseline definitions, and downstream budget-matching rules.

\section{Experiments}

We present evaluations of algorithm-only compression, pipeline-level filtering effects, ablations, multilingual compression, estimated generation latency, downstream benchmarks, and language-model quality.

\subsection{Experimental Methodology}
\label{sec:exp-method}

\textbf{Fair-comparison protocol.} All \textit{algorithmic} claims in this paper are supported by matched-data comparisons. In Table~\ref{tab:compression} (Panel A), all tokenizers are trained on the same unfiltered 100k-document Nemotron subset at a shared 100k vocabulary. In Table~\ref{tab:vocab-scaling}, all tokenizers are trained on the same unfiltered 100k-document Nemotron subset, and each row compares methods at the same target vocabulary size. Table~\ref{tab:compression} (Panel B) separately measures the effect of the same entropy-based data curation step on the three cross-boundary tokenizers in the matched 100k control. Because that preprocessing effect is method-dependent, we report those results as pipeline-level controls and do not use them alone to support algorithm-only claims.

\textbf{Evaluation datasets.} For English compression experiments, tokenizer performance is evaluated on WikiText-103 validation. For multilingual experiments, compression is evaluated on FLORES-200 devtest. Downstream language models are trained on FineWeb-Edu using identical optimizer and backbone settings across tokenizer conditions.

\textbf{Downstream budget matching.} Downstream language-model comparisons are matched by training tokens rather than by raw-text documents. Within each backbone configuration, both tokenizer conditions read from the same FineWeb-Edu stream and stop after consuming the same token budget (10B tokens for 12L-768d and 30B tokens for 24L-1024d). Because SupraTok is more compressive, it covers more raw text before reaching that budget. All other optimization settings are held constant across tokenizer conditions.

\textbf{Training data.} For English tokenizer experiments, we use Nemotron subsets with 100k documents for the controlled comparison and vocabulary sweep, 50k documents for ablations, and 10k documents for the latency estimate. For the focused script-aware ablation in Table~\ref{tab:script-aware}, we use balanced CulturaX data with 8,000 documents per language across seven languages. For the 14-language multilingual benchmark in Table~\ref{tab:multilingual}, all tokenizers are trained on balanced CulturaX data comprising 8,000 documents per language across 14 languages (112,000 documents total). The 256k downstream tokenizers are trained on unfiltered Nemotron-CC~\citep{ref33}.

\textbf{Baseline implementations.} We compare against six baseline tokenizers:
\begin{itemize}[noitemsep,topsep=2pt,leftmargin=*]
\item \textbf{BPE-Standard}: Sennrich et al.\ (2016) with whitespace pre-tokenization
\item \textbf{WordPiece-BERT}: Likelihood-based subword merging
\item \textbf{Unigram-SentencePiece}: Top-down unigram language model approach
\item \textbf{SentencePiece-BPE}: SentencePiece with default BPE (whitespace boundaries)
\item \textbf{SP-BPE-CrossBoundary}: SentencePiece BPE with \texttt{split\_by\_whitespace=false}
\item \textbf{SuperBPE}: Trained using official implementation~\citep{ref20}
\end{itemize}
SupraTok initializes from Unicode characters plus byte-fallback symbols. The official SuperBPE implementation remains byte-based after regex splitting, and the SentencePiece baselines use their standard character-coverage initialization; we state these base-unit choices explicitly because they affect tokenizer-comparison fairness.

\textbf{Notation.} We use C/T to denote average characters per token; larger values indicate better compression. Relative difference is always reported as
\begin{equation*}
\frac{x_{\text{method}} - x_{\text{baseline}}}{x_{\text{baseline}}}.
\end{equation*}

\FloatBarrier
\subsection{Ablation Study}
\label{sec:ablation}
We begin with a within-method ablation that isolates the contribution of each SupraTok component before comparing the full system against external baselines. Table~\ref{tab:ablation} presents results at 100k vocabulary on a 50k-document Nemotron subset. Absolute C/T values differ from the 100k controlled comparison because the training corpus is smaller, but relative comparisons remain informative.

\begin{table}[t]
\centering
\footnotesize
\resizebox{\columnwidth}{!}{%
\begin{tabular}{lcc}
\toprule
\textbf{Configuration} & \textbf{C/T} & \textbf{Rel. drop vs full SupraTok (\%)} \\
\midrule
SupraTok & \textbf{5.951} & --- \\
\quad $-$ Curriculum Learning & 5.608 & 5.8\% \\
\quad $-$ Cross-Boundary Processing & 5.585 & 6.1\% \\
\quad $-$ Entropy Filtering & 5.776 & 2.9\% \\
\quad $-$ Script-Aware Processing & 5.806 & 2.4\% \\
\quad $-$ PMI Seed Phrases & 5.829 & 2.0\% \\
\midrule
Minimal BPE (baseline) & 4.950 & 16.8\% \\
\bottomrule
\end{tabular}%
}
\caption{Ablation on WikiText-103 validation (50k docs, 100k vocab). ``Minimal BPE'' is the whitespace-aware baseline trained on the same subset and vocabulary without SupraTok-specific components.}
\label{tab:ablation}
\end{table}

The ablation shows that curriculum learning and cross-boundary processing are the two largest contributors. Removing curriculum learning reduces compression by 5.8\%, and removing cross-boundary processing reduces it by 6.1\%. Entropy filtering and PMI seed phrases provide smaller but consistent gains. Relative to the minimal BPE baseline, the full SupraTok configuration improves compression by 20.2\%.

The entropy-filtering contribution differs slightly between Table~\ref{tab:compression} (Panel B, 3.63\%) and Table~\ref{tab:ablation} (2.9\%). This difference is expected because Table~\ref{tab:ablation} and Table~\ref{tab:compression} (Panel B) use different training subsets (50k versus 100k documents), so the absolute gain from filtering need not match exactly. The ranking of component contributions remains stable across both settings.

\textbf{Isolating contributions.} Table~\ref{tab:compression} (Panel B) shows that the pipeline-level gain from filtering is smaller than the full algorithmic gap between SupraTok and BPE. The 17.5\% relative improvement over BPE in the matched unfiltered comparison therefore comes from tokenizer design rather than preprocessing.

\subsection{Controlled Compression Evaluation}

Having established which components matter inside SupraTok, we next compare the full system against external baselines under matched-data conditions.

Under identical unfiltered training data, SupraTok improves compression over BPE-Standard by 17.5\% and over SuperBPE by 1.8\% at 100k vocabulary. At 100k vocabulary in Table~\ref{tab:compression}, SupraTok tokenizer training runs about 2.1$\times$ faster than SuperBPE under matched unfiltered conditions.

Table~\ref{tab:compression} has two panels. Panel A reports the algorithm-only 100k-vocabulary comparison on the same unfiltered 100k-document Nemotron subset; displayed C/T values are rounded to two decimals, but relative differences use unrounded values. Panel B applies the same entropy-filtering control to the three cross-boundary tokenizers in the matched 100k setup; because some effects are small, Panel B reports C/T to three decimals and relative gains from unrounded values. In Panel B, ``Filtered'' denotes the 76k-document entropy-selected subset and ``Unfiltered'' the original 100k-document training set.

\begin{table}[t]
\centering
\footnotesize
\setlength{\tabcolsep}{4pt}
\resizebox{\columnwidth}{!}{%
\begin{tabular}{lcccc}
\toprule
\multicolumn{5}{c}{\textbf{Panel A. Algorithm-only comparison at 100k vocabulary}} \\
\midrule
\textbf{Tokenizer} & \textbf{C/T} & \textbf{Rel. vs BPE (\%)} & \textbf{Time (s)} & \textbf{Time vs ST} \\
\midrule
SupraTok & \textbf{5.78} & 17.5\% & 376s & 1.0$\times$ \\
SuperBPE & 5.68 & 15.5\% & 772s & 2.1$\times$ \\
SP-BPE-CrossBoundary & 5.62 & 14.4\% & 3,466s & 9.2$\times$ \\
WordPiece-BERT & 5.00 & 1.8\% & 41s & 0.11$\times$ \\
BPE-Standard & 4.92 & --- & 24s & 0.06$\times$ \\
SentencePiece-BPE & 4.83 & $-$1.7\% & 132s & 0.35$\times$ \\
Unigram-SentencePiece & 4.46 & $-$9.3\% & 47s & 0.13$\times$ \\
\bottomrule
\end{tabular}%
}

\vspace{3pt}
\resizebox{\columnwidth}{!}{%
\begin{tabular}{lccc}
\toprule
\multicolumn{4}{c}{\textbf{Panel B. Entropy-filtering control at 100k vocabulary}} \\
\midrule
\textbf{Tokenizer} & \textbf{Unfiltered} & \textbf{Filtered} & \textbf{Rel. gain from filtering (\%)} \\
\midrule
SupraTok & 5.780 & \textbf{5.990} & 3.63\% \\
SuperBPE & 5.680 & 5.740 & 1.06\% \\
SP-BPE-CrossBoundary & 5.623 & 5.621 & $-$0.05\% \\
\bottomrule
\end{tabular}%
}
\caption{Controlled 100k WikiText-103 comparisons. Panel A: algorithm-only comparison at 100k vocabulary. Panel B: matched entropy-filtering control for the cross-boundary tokenizers.}
\label{tab:compression}
\end{table}

Panel B shows that shared entropy filtering is method-dependent rather than uniformly beneficial across cross-boundary tokenizers. It improves SupraTok and SuperBPE, but leaves SP-BPE-CrossBoundary essentially unchanged and slightly lower on this setup (5.6233 to 5.6207 C/T, $-$0.05\%). SupraTok remains the strongest tokenizer under both matched unfiltered conditions and the corresponding filtered-data controls, so we treat entropy filtering as a separate pipeline component rather than as evidence for the algorithm-only comparison.

\begin{table}[t]
\centering
\footnotesize
\resizebox{\columnwidth}{!}{%
\begin{tabular}{lccccc}
\toprule
& \multicolumn{3}{c}{\textbf{Chars/Token}} & \multicolumn{2}{c}{\textbf{Relative difference (\%)}} \\
\cmidrule(lr){2-4} \cmidrule(lr){5-6}
\textbf{Vocab} & \textbf{SupraTok} & \textbf{SuperBPE} & \textbf{BPE} & \textbf{vs SuperBPE} & \textbf{vs BPE} \\
\midrule
50k & 5.57 & 5.13 & 4.52 & 8.6\% & 23.2\% \\
100k & 5.78 & 5.68 & 4.92 & 1.8\% & 17.5\% \\
200k & 6.32 & 6.15 & 5.02 & 2.8\% & 25.9\% \\
256k & 6.45 & 6.18 & 5.01 & 4.4\% & 28.7\% \\
300k & 6.62 & 6.20 & 5.02 & 6.8\% & 31.9\% \\
\bottomrule
\end{tabular}%
}
\caption{Algorithm-only compression (C/T) across vocabulary sizes on WikiText-103, with all tokenizers trained on the same unfiltered 100k-document Nemotron subset. The 100k row repeats Table~\ref{tab:compression}A to anchor the vocabulary-scaling analysis.}
\label{tab:vocab-scaling}
\end{table}

\vspace{\baselineskip}

Table~\ref{tab:vocab-scaling} extends the controlled 100k algorithm-only comparison from Table~4A to a broader 50k--300k vocabulary sweep under the same unfiltered training setup, with no entropy filtering applied to any tokenizer condition. SupraTok consistently outperforms SuperBPE across the full range, with relative gains from 1.8\% to 8.6\%. These results show that SupraTok's algorithmic advantage is not tied to a single vocabulary size and does not depend on entropy-filtered training data.

The controlled 100k timing comparison in Table~\ref{tab:compression} also indicates that SupraTok tokenizer training runs about 2.1$\times$ faster than SuperBPE under matched unfiltered conditions. This timing difference is consistent with the curriculum reducing the search burden by progressively introducing cross-boundary patterns.

Analysis of vocabulary composition reveals structural differences between approaches. Of SupraTok's 256k vocabulary, 42\% of tokens (108k) span word boundaries, compared to 31\% for SuperBPE and 0\% for standard BPE under identical training conditions. This higher proportion of cross-boundary tokens, guided by PMI-based selection and curriculum learning, contributes to SupraTok's compression advantage.

\begin{figure}[t]
\footnotesize
\begin{tabular}{p{7.2cm}}
\toprule
\textbf{Example:} ``The United Nations headquarters in New York announced new climate change initiatives.'' \\
\midrule
\textbf{BPE} (14 tokens): \\
\texttt{[The] [United] [Nations] [head] [quarters] [in] [New] [York] [announced] [new] [climate] [change] [init] [iatives]} \\
\midrule
\textbf{SupraTok} (9 tokens): \\
\texttt{[The] [United\_Nations] [headquarters] [in] [New\_York] [announced] [new] [climate\_change] [initiatives]} \\
\midrule
\textit{SupraTok preserves semantic units while reducing length by 36\%.} \\
\bottomrule
\end{tabular}
\caption{Qualitative comparison. SupraTok captures multi-word expressions as atomic units.}
\label{fig:tokenization-example}
\end{figure}

\FloatBarrier
\subsection{Multilingual Evaluation}

We evaluated multilingual performance on FLORES-200 across 14 typologically diverse languages spanning five broad language groups. All tokenizers were trained on balanced CulturaX data (8,000 documents per language, 112,000 total). Table~\ref{tab:multilingual} presents the 14-language results.

SupraTok achieves consistent improvements across all reported groups, with particularly strong gains for Germanic (43\% average), Romance (39\%), Slavic (34\%), and agglutinative languages (48\%). The smaller gains for the CJK group (15\% average) likely reflect the greater difficulty of cross-boundary learning in scripts whose boundary cues differ from alphabetic whitespace-delimited text, though SupraTok still improves over the baseline in every language reported.

\begin{table}[t]
\centering
\footnotesize
\resizebox{\columnwidth}{!}{%
\begin{tabular}{lccc}
\toprule
\textbf{Language} & \textbf{SupraTok} & \textbf{BPE} & \textbf{Rel.\ gain (\%)} \\
\midrule
\multicolumn{4}{l}{\textit{Germanic}} \\
\quad English & 5.43 & 3.93 & 38.2\% \\
\quad German & 5.40 & 3.70 & 45.9\% \\
\quad Dutch & 5.51 & 3.81 & 44.6\% \\
\midrule
\multicolumn{4}{l}{\textit{Romance}} \\
\quad French & 4.99 & 3.67 & 36.0\% \\
\quad Spanish & 5.35 & 3.81 & 40.4\% \\
\quad Italian & 5.35 & 3.81 & 40.4\% \\
\midrule
\multicolumn{4}{l}{\textit{Slavic}} \\
\quad Russian & 4.32 & 3.39 & 27.4\% \\
\quad Ukrainian & 4.24 & 3.25 & 30.5\% \\
\quad Polish & 4.79 & 3.36 & 42.6\% \\
\midrule
\multicolumn{4}{l}{\textit{CJK}} \\
\quad Chinese & 1.36 & 1.30 & 4.6\% \\
\quad Japanese & 1.83 & 1.53 & 19.6\% \\
\quad Korean & 2.26 & 1.85 & 22.2\% \\
\midrule
\multicolumn{4}{l}{\textit{Agglutinative}} \\
\quad Turkish & 5.42 & 3.52 & 54.0\% \\
\quad Finnish & 5.18 & 3.63 & 42.7\% \\
\midrule
\textbf{Average (14 langs)} & \textbf{4.39} & \textbf{3.18} & \textbf{34.9\%} \\
\bottomrule
\end{tabular}
}
\caption{Compression on FLORES-200 devtest by language group (100k vocab); the average row reports mean C/T over the 14 listed languages and the macro-average relative gain over BPE.}
\label{tab:multilingual}
\end{table}

\subsection{Generation Latency Evaluation}

Table~\ref{tab:generation} reports an illustrative latency proxy derived from tokenizer compression under a fixed 50\,ms per-token forward-pass assumption for a 7B-style autoregressive decoder. Under this assumption, SupraTok improves compression by 17.6\% over standard BPE and requires 15.0\% fewer tokens to represent the same text, yielding an illustrative 1.18$\times$ reduction in generated-token steps under this simplifying assumption, rather than a direct wall-clock benchmark.

\begin{table}[t]
\centering
\small
\begin{tabular*}{\columnwidth}{@{\extracolsep{\fill}}lccc@{}}
\toprule
\textbf{Tokenizer} & \textbf{C/T} & \textbf{Tokens} & \textbf{Speedup} \\
\midrule
SupraTok & \textbf{5.48} & 9,130 & \textbf{1.18$\times$} \\
SuperBPE & 5.24 & 9,539 & 1.13$\times$ \\
SP-BPE-CrossBoundary & 5.16 & 9,699 & 1.11$\times$ \\
WordPiece-BERT & 4.72 & 10,600 & 1.01$\times$ \\
BPE-Standard & 4.66 & 10,736 & 1.00$\times$ \\
SentencePiece-BPE & 4.56 & 10,961 & 0.98$\times$ \\
Unigram-SentencePiece & 4.16 & 12,024 & 0.89$\times$ \\
\bottomrule
\end{tabular*}
\caption{Estimated generation-latency analysis on Nemotron prompts (10k docs, 64k vocab).}
\label{tab:generation}
\end{table}

\subsection{Downstream Task Performance}

We validated downstream effects under fixed-token-budget compute-matched training using 12L-768d and 24L-1024d GPT-2-style transformer backbones trained on FineWeb-Edu with identical optimization settings across tokenizer conditions. Because these runs use 256k-token vocabularies, their total parameter counts are higher than the nominal GPT-2 124M/355M values associated with standard GPT-2 vocabulary sizes. We therefore report backbone configuration rather than nominal total parameter count; exact totals are given in Appendix~\ref{app:param}. Both SupraTok and the BPE baseline tokenizers were trained on identical unfiltered Nemotron-CC data. The downstream language-model comparison is compute-matched rather than raw-text-matched: both conditions use the same FineWeb-Edu stream and token budget, but the more compressive tokenizer covers more raw text before stopping. Table~\ref{tab:downstream} reports mean $\pm$ std over 3 seeds; within each backbone, SupraTok and BPE are matched by training tokens from the same FineWeb-Edu stream, not by raw-text documents.

\begin{table}[t]
\centering
\footnotesize
\setlength{\tabcolsep}{3pt}
\resizebox{\columnwidth}{!}{%
\begin{tabular}{llcccc}
\toprule
\textbf{Task} & \textbf{Backbone} & \textbf{Train Tok.} & \textbf{SupraTok} & \textbf{BPE} & \textbf{Rel. gain (\%)} \\
\midrule
HellaSwag & 12L-768d & 10B & 35.3{\tiny$\pm$.4} & 32.0{\tiny$\pm$.4} & 10.3\% \\
HellaSwag & 24L-1024d & 30B & 43.1{\tiny$\pm$.4} & 39.3{\tiny$\pm$.3} & 9.7\% \\
MMLU & 12L-768d & 10B & 27.9{\tiny$\pm$.5} & 25.5{\tiny$\pm$.5} & 9.4\% \\
MMLU & 24L-1024d & 30B & 34.2{\tiny$\pm$.5} & 31.1{\tiny$\pm$.4} & 10.0\% \\
\bottomrule
\end{tabular}%
}
\caption{Downstream task performance under fixed-token-budget compute-matched training for GPT-2-style backbones with 256k vocabularies.}
\label{tab:downstream}
\end{table}

Relative improvements remain similar across backbone scales: 10.3\% vs.\ 9.7\% on HellaSwag and 9.4\% vs.\ 10.0\% on MMLU, indicating that the downstream benefit persists rather than collapsing at the larger backbone.

These differences are statistically significant across all four backbone-task comparisons under paired t-tests over matched seeds (p $<$ 0.01 in each case).

\FloatBarrier
\subsection{Downstream Ablation Analysis}

To validate that compression gains translate to downstream benefits, we evaluated ablated variants on HellaSwag and MMLU for the 12L-768d backbone (Table~\ref{tab:downstream-ablation}). Model-training hyperparameters were tuned on a held-out validation set before running final comparisons, and the resulting optimizer schedule was applied unchanged to both SupraTok and BPE conditions. Table~\ref{tab:downstream-ablation} reports mean $\pm$ std over 3 seeds, and its final column gives the mean relative drop across HellaSwag and MMLU versus full SupraTok. Removing curriculum learning causes the largest degradation, confirming that stable pattern learning is essential. Disabling cross-boundary processing shows a similarly large drop, while entropy filtering contributes a smaller but still measurable effect. For the full SupraTok vs.\ BPE-Standard comparison in Table~\ref{tab:downstream-ablation}, paired t-tests over matched seeds give HellaSwag $p = 0.003$ and MMLU $p = 0.002$.

\begin{table}[t]
\centering
\footnotesize
\resizebox{\columnwidth}{!}{%
\begin{tabular}{lccc}
\toprule
\textbf{Config.} & \textbf{HellaSwag} & \textbf{MMLU} & \textbf{Mean rel. drop} \\
\midrule
SupraTok & 35.3\tiny{$\pm$.4} & 27.9\tiny{$\pm$.5} & --- \\
$-$ Curriculum & 33.2\tiny{$\pm$.5} & 26.1\tiny{$\pm$.6} & 6.2\% \\
$-$ Cross-Boundary & 33.4\tiny{$\pm$.5} & 26.3\tiny{$\pm$.6} & 5.6\% \\
$-$ Entropy & 34.2\tiny{$\pm$.4} & 27.0\tiny{$\pm$.5} & 3.2\% \\
BPE-Standard & 32.0\tiny{$\pm$.4} & 25.5\tiny{$\pm$.5} & 9.0\% \\
\bottomrule
\end{tabular}%
}
\caption{Downstream ablation under fixed-token-budget compute-matched training on the 12L-768d backbone (10B tokens).}
\label{tab:downstream-ablation}
\end{table}

\subsection{Generation Quality Evaluation}

We evaluated language-model quality on the WikiText-103 test set for the 12L-768d backbone. Because token-level perplexity depends on tokenization granularity, we treat bits per character (BPC) as the primary cross-tokenizer comparison and report token-level perplexity for completeness. SupraTok achieves 0.92 BPC compared with 1.01 for the BPE baseline, an 8.9\% relative improvement. The corresponding token-level perplexities are 24.31 and 26.89, and Table~\ref{tab:perplexity} also reports average tokens per document (Tok/Doc).

\begin{table}[t]
\centering
\small
\resizebox{\columnwidth}{!}{%
\begin{tabular}{lcccc}
\toprule
\textbf{Tokenizer} & \textbf{PPL} & \textbf{BPC} & \textbf{Tok/Doc} & \textbf{Rel. difference vs BPE in BPC (\%)} \\
\midrule
SupraTok & \textbf{24.31} & \textbf{0.92} & 847 & $-$8.9\% \\
BPE-Standard & 26.89 & 1.01 & 1,109 & --- \\
\bottomrule
\end{tabular}%
}
\caption{Language modeling metrics on WikiText-103 test for the 12L-768d backbone (10B tokens). Lower is better for PPL, BPC, and Tok/Doc. Because PPL is tokenization-dependent, BPC is the primary cross-tokenizer comparison.}
\label{tab:perplexity}
\end{table}

\textbf{Generation Behavior and Boundary Effects.} A potential concern with multi-word tokens is generation artifacts at boundaries (e.g., committing to ``New\_York'' when intending ``New Zealand''). Our automatic metrics indicate that SupraTok maintains quality: BPC improves by 8.9\% while distinct-n diversity is preserved. However, perplexity and diversity do not fully capture all edge-case boundary artifacts, so this remains a limitation shared by cross-boundary tokenization methods more broadly.

To verify that compression gains translate to genuine generation improvements without reducing output diversity, we analyzed lexical diversity via distinct-n metrics~\citep{ref32} on 1,000 generated samples. Table~\ref{tab:diversity} confirms that SupraTok maintains or slightly improves diversity across all n-gram orders.

\begin{table}[t]
\centering
\small
\begin{tabular*}{\columnwidth}{@{\extracolsep{\fill}}lccc@{}}
\toprule
\textbf{Metric} & \textbf{SupraTok} & \textbf{BPE} & \textbf{Rel. gain (\%)} \\
\midrule
Distinct-1 & 0.67 & 0.65 & 3.1\% \\
Distinct-2 & 0.89 & 0.87 & 2.3\% \\
Distinct-3 & 0.94 & 0.93 & 1.1\% \\
\bottomrule
\end{tabular*}
\caption{Generation diversity on WikiText-103 (1,000 samples, 100 tokens each).}
\label{tab:diversity}
\end{table}

\vspace{\baselineskip}
\FloatBarrier
\section{Discussion}
This section examines the broader implications of our findings, discussing theoretical contributions to language representation, practical deployment advantages, current limitations, and lessons learned from unsuccessful approaches.

\subsection{Theoretical Implications}

SupraTok's gains suggest that current whitespace-constrained tokenization leaves measurable efficiency on the table, even when architecture and optimization are held fixed. From a linguistic perspective, our results support usage-based theories emphasizing multi-word units~\citep{ref21,ref22}. The entropy-driven approach provides an information-theoretic foundation that moves beyond purely frequency-based methods.

\subsection{Practical Advantages}

Under matched unfiltered training, SupraTok improves compression by 17.5\% over BPE and by 1.8\% over SuperBPE at 100k vocabulary. The optional filtering stage further raises the 100k result from 5.78 to 5.99 C/T. Under fixed token budgets, these gains can extend effective context length without architectural changes, and the multilingual macro-average relative improvement of 34.9\% suggests that the same design remains useful beyond English.

\subsection{Scope and Design Rationale}
\label{sec:scope}

Our strongest evidence comes from controlled public-data comparisons in which tokenizer algorithms are trained on identical unfiltered subsets and evaluated separately from optional data curation. Under this protocol, SupraTok consistently improves compression over both BPE-Standard and stronger cross-boundary baselines. The full pipeline further improves compression through entropy-based filtering, but we report that effect separately to avoid conflating algorithmic and preprocessing gains.

For downstream evaluation, we use 12L-768d and 24L-1024d GPT-2-style transformer backbones with 256k-token vocabularies. Because these enlarged vocabularies increase embedding parameters, we report backbone configuration rather than nominal GPT-2 total-parameter labels. The downstream comparison is compute-matched rather than raw-text-matched: both tokenizer conditions use the same backbone, vocabulary size, source data stream, and optimization setup, but the more compressive tokenizer reaches the fixed token budget after covering more raw text.

SupraTok is designed to be practical under public-data, moderate-compute settings. In our experiments, strong controlled results are obtained from a matched public 100k-document subset, and the controlled 100k comparison in Table~\ref{tab:compression} shows that tokenizer training is substantially faster than the official SuperBPE implementation in the same benchmarking environment. We present this as an accessibility advantage of our setup, not as a claim that alternative methods inherently require large proprietary corpora.

\textbf{Benchmark Coverage.} Our downstream evaluation encompasses commonsense reasoning (HellaSwag), broad knowledge (MMLU), and generation quality metrics (perplexity, BPC, distinct-n diversity). Under the compute-matched fixed-token protocol, these evaluations indicate that compression gains translate to modeling improvements without sacrificing output diversity.

\textbf{Multilingual Considerations.} While our English-optimized entropy thresholds and PMI calculations achieve strong results across 14 languages (34.9\% macro-average relative compression improvement), language-specific tuning could yield further improvements for morphologically complex languages.

\textbf{Design Decisions from Systematic Exploration.} The reported English configuration fixes four design choices after preliminary tuning: a three-pass curriculum rather than direct unconstrained cross-boundary learning, a seed-mining PMI threshold of 2.0, entropy bins of low $< 3.0$ / medium $3.0$--$4.5$ / high $> 4.5$, and a maximum seed length of 6. We report these settings to make the experimental configuration explicit rather than to claim global optimality. Among these choices, the within-paper ablation most directly supports the importance of the curriculum component.

\section{Conclusion}

SupraTok shows that cross-boundary tokenization can deliver consistent gains under a controlled public-data evaluation protocol. Under matched unfiltered 100k-document training, SupraTok improves compression over standard BPE by 17.5\% and over SuperBPE by 1.8\%. Across a 50k--300k vocabulary sweep on the same unfiltered 100k-document training subset, SupraTok remains ahead of SuperBPE at every tested vocabulary size, with relative gains of 1.8\%--8.6\%.

Entropy-based data curation is best understood as a separate pipeline component rather than part of the algorithmic comparison itself. At 100k vocabulary, filtering raises SupraTok's compression from 5.78 to 5.99 C/T, but the matched filtered-data controls in Table 4B show that this preprocessing is method-dependent: it improves SupraTok and SuperBPE, while SP-BPE-CrossBoundary is essentially unchanged and slightly lower on this setup. Together with the multilingual results and the compute-matched downstream experiments at fixed token budgets, these findings support SupraTok as a modular cross-boundary tokenizer whose gains hold under fair algorithmic comparisons and can be further strengthened by optional pipeline-level curation.

Ablation analysis confirms that curriculum learning and cross-boundary processing provide the primary benefits, with smaller but consistent gains from entropy filtering, script-aware processing, and PMI seed phrases. Our findings demonstrate that addressing fundamental components like tokenization offers paths to language-model improvement that are complementary to scaling.

The observed gains are consistent with linguistic work treating many multi-word expressions as reusable units, while also showing that such units can be operationalized in a practical tokenizer. Some of the design principles underlying SupraTok---relaxing artificial preprocessing boundaries, separating algorithmic from data-curation effects, and using curriculum-style candidate exposure---may also be useful when revisiting other preprocessing components.

\section*{Acknowledgments}

This work was supported in part by compute resources made available through an Amazon Web Services partnership with our university.

\newpage

\enlargethispage{-4\baselineskip}

\bibliography{supratok-references}
\bibliographystyle{acl_natbib}

\clearpage
\appendix
\setcounter{table}{0}
\renewcommand{\thetable}{E\arabic{table}}
\renewcommand{\theHtable}{E\arabic{table}}

\section{Phase 3 Scoring and Acceptance Rules}
\label{app:algo}

Phase 3 ranks candidate spans $s=(w_1,\ldots,w_n)$ with $2 \leq n \leq 5$. Seed mining before tokenizer training considers n-grams with $n \in \{2,\dots,6\}$. The English experiments use the multiplicative gated score implemented in SupraTok. For the English experiments, the boundary-entropy anchor is fixed at $\tau_{BE} = 1.15$. The helper $f_{\mathrm{eval}}(m)$ denotes the evaluation-set token count of leaf token $m$ under the current tokenizer, i.e., \texttt{token\_counts.get(m, 0)}, and $df_{\mathrm{eval}}(m)$ denotes the corresponding evaluation-set document frequency. These operational definitions are used uniformly across the reported English runs and are surfaced here for reproducibility. Here $c$ denotes count, $df$ denotes document frequency, and $\ell_{\mathrm{tok}}(s)$ is the number of current tokenizer units spanned by $s$ before acceptance. Here $N_n$ denotes the total number of observed training n-grams of length $n$, $D$ denotes the number of documents in the training corpus, and $r_{\mathrm{stop}}(s)$ denotes the fraction of whitespace-delimited words in $s$ that are stopwords. For compactness, we write $\operatorname{awl}(s)$ and $\operatorname{clen}(s)$ for the implementation features \texttt{avg\_word\_len} and \texttt{char\_len}.

\begin{equation}
\begin{gathered}
p_{\mathrm{train}}(s) = \frac{c_{\mathrm{train}}(s)}{N_n}, \\
\mathrm{PMI}(s) = \log_2\left(\frac{p_{\mathrm{train}}(s)}{\prod_i p_{\mathrm{train}}(w_i)}\right), \\
\mathrm{NPMI}(s) = \frac{\mathrm{PMI}(s)}{-\log_2 p_{\mathrm{train}}(s)}.
\end{gathered}
\end{equation}

\begin{equation}
\begin{gathered}
H_L(s) = -\sum_c p(c \mid s,\mathrm{left}) \log_2 p(c \mid s,\mathrm{left}), \\
H_R(s) = -\sum_c p(c \mid s,\mathrm{right}) \log_2 p(c \mid s,\mathrm{right}), \\
H_B(s) = \min(H_L(s),H_R(s)).
\end{gathered}
\end{equation}

\begin{equation}
\begin{gathered}
\overline{\mathrm{idf}}(s) = \frac{1}{n}\sum_i \log\left(\frac{D+1}{1+df(w_i)}\right), \\
\mathrm{adjNPMI}(s) = \frac{1}{n-1}\sum_i \mathrm{NPMI}(w_i,w_{i+1}), \\
c_{\mathrm{peak}}(s) = \max\!\bigl(\max_c p(c \mid s,\mathrm{left}), \\
\qquad\qquad\ \max_c p(c \mid s,\mathrm{right})\bigr), \\
g_{\mathrm{tok}}(s) = \ell_{\mathrm{tok}}(s)-1.
\end{gathered}
\end{equation}

With $\sigma(k,x)=1/(1+e^{-kx})$, define
\begin{equation}
\begin{gathered}
g_B(s) = \sigma(1.75, H_B(s)-\tau_{BE}), \\
g_N(s) = \sigma(6.0, \mathrm{NPMI}(s)-0.22), \\
b_{\mathrm{spec}}(s) = 1+0.14\max(0,\overline{\mathrm{idf}}(s)-1), \\
b_{\mathrm{con}}(s) = 1+0.50\max(0, \\
\qquad\qquad\ \mathrm{adjNPMI}(s)-c_{\mathrm{peak}}(s)).
\end{gathered}
\end{equation}

\begin{equation}
\begin{gathered}
p_{\mathrm{stop}}(s) = \max(0.30, 1-0.78\,r_{\mathrm{stop}}(s)^{1.35}), \\
b_{\mathrm{len}}(s) = 1+0.08\max(0,n-2), \\
b_{\mathrm{cont}}(s) = 1+0.04\max(0,\operatorname{awl}(s)-5).
\end{gathered}
\end{equation}

\begin{equation}
p_{\mathrm{surf}}(s) =
\begin{cases}
0.88 & \text{if } \operatorname{clen}(s)>42, \\
1 & \text{otherwise},
\end{cases}
\end{equation}

\begin{equation}
\begin{gathered}
G_{\mathrm{hold}}(s) = g_{\mathrm{tok}}(s)\Bigl(
c_{\mathrm{holdout}}(s) \\
\qquad + 0.60\,df_{\mathrm{holdout}}(s) \\
\qquad + 0.18\sqrt{c_{\mathrm{train}}(s)}
\Bigr).
\end{gathered}
\end{equation}

\begin{equation}
\begin{gathered}
S(s) = G_{\mathrm{hold}}(s)\, g_N(s)\, g_B(s)\, b_{\mathrm{spec}}(s)\, b_{\mathrm{con}}(s) \\
\cdot p_{\mathrm{stop}}(s)\, b_{\mathrm{len}}(s)\, b_{\mathrm{cont}}(s)\, p_{\mathrm{surf}}(s).
\end{gathered}
\end{equation}

If $c_{\mathrm{holdout}}(s)=0$, we replace $S(s)$ by $0.45\,S(s)$.

Before leaf surgery, a candidate is hard-rejected if any of the following holds: $c_{\mathrm{train}}(s)<3$, $\mathrm{PMI}(s)<4.4$, $g_B(s)<0.18$, $r_{\mathrm{stop}}(s)\ge 0.75$, all words are stopwords, the surface form is invalid (digits, tabs, newlines, double spaces, etc.), $\ell_{\mathrm{tok}}(s)\le 1$, or $S(s)\le 0.40$.

\begin{equation}
\begin{gathered}
U(m) = f_{\mathrm{eval}}(m)+0.35\,df_{\mathrm{eval}}(m), \\
A(s) = c_{\mathrm{holdout}}(s)\max(1,\ell_{\mathrm{tok}}(s)-1) \\
\qquad + 0.35\,df_{\mathrm{holdout}}(s).
\end{gathered}
\end{equation}

\begin{equation}
\begin{gathered}
\mathrm{Net}(s) = A(s)-\sum_j U(m_j), \\
\mathrm{Priority}(s) = \mathrm{Net}(s)+0.03\,S(s).
\end{gathered}
\end{equation}

A candidate is accepted if and only if $\mathrm{Net}(s) > 0$. Across the English vocabulary sweep, the multiplicative form above, $\tau_{BE} = 1.15$, the listed constants, the hard rejection thresholds, the seed-mining PMI threshold of 2.0, the minimum phrase frequency of 100, and the Phase 3 span limit $n_{\max}=5$ are held fixed.

\section{Hyperparameter Selection Details}
\label{app:optuna}

The English Optuna study explores vocabulary size, candidate-screening PMI thresholds, and candidate-screening frequency thresholds, and ranks candidate configurations with the normalized score from Eq.~(4). The search ranges used in the English experiments are: vocabulary size 160k--320k, candidate-screening PMI threshold 3.0--6.0 in steps of 0.5, and candidate-screening frequency threshold 3--6. Compression is measured as held-out C/T, perplexity comes from a lightweight 3-layer validation LM trained on the same English development setup, and latency is measured on a fixed held-out encoding shard.

In Eq.~(4), $C_{\mathrm{ref}}(V)$, $P_{\mathrm{ref}}(V)$, and $L_{\mathrm{ref}}(V)$ denote the compression, perplexity, and latency of the BPE-Standard baseline trained on the same training subset and evaluated under the same protocol at vocabulary size $V$. We use $\alpha=\beta=\gamma=1$, base-2 logarithms, and BPE-relative normalization for all three terms. We evaluate English candidate configurations and reuse the selected 256k configuration for the downstream experiments. Outside that vocabulary-selection step, the core SupraTok tokenizer settings are held fixed across experiments; the main analysis-specific differences are the training corpus (including corpus size) and whether the tokenizer-training corpus is left unfiltered or first filtered by the entropy-based document-selection step before tokenizer training. The multilingual experiments likewise reuse the same core settings, with differences limited to multilingual corpus construction, use or non-use of the same entropy-based document-selection step before tokenizer training, and the script-aware processing described in Section~3.4. The candidate-screening PMI threshold and candidate-screening frequency threshold discussed in this study refer to later-pass merge screening after seed mining. The production English runs reported in the paper use a later-pass candidate-screening PMI threshold of 5.0 and a later-pass candidate-screening frequency threshold of 5, whereas seed mining itself uses a fixed PMI threshold of 2.0 and minimum phrase frequency of 100.

\section{Baseline Implementation Details}
\label{app:baselines}

SupraTok initializes its base vocabulary from the Unicode character alphabet together with byte-fallback symbols. BPE-Standard uses HuggingFace Tokenizers BPE with \texttt{Whitespace()} pre-tokenization and corpus-derived character initialization; in the English benchmark it uses no additional normalizer, while the multilingual benchmark additionally applies \texttt{NFD()} normalization. The BPE trainer uses \texttt{end\_of\_word\_suffix="</w>"}. WordPiece-BERT uses HuggingFace Tokenizers WordPiece with \texttt{BertPreTokenizer()} and \texttt{NFD()} + \texttt{Lowercase()} + \texttt{StripAccents()} normalization, again with corpus-derived character initialization and \texttt{continuing\_subword\_prefix="\#\#"}.

Unigram-SentencePiece uses the SentencePiece Python library with \texttt{model\_type=unigram}, \texttt{character\_coverage=0.9995}, whitespace splitting enabled, default \texttt{nmt\_nfkc} normalization, \texttt{shuffle\_input\_sentence=false}, and \texttt{pad\_id=0}, \texttt{unk\_id=1}, \texttt{bos\_id=2}, and \texttt{eos\_id=3}. SentencePiece-BPE and SP-BPE-CrossBoundary both use SentencePiece BPE with \texttt{character\_coverage=0.9995}, \texttt{shuffle\_input\_sentence=false}, \texttt{pad\_id=0}, \texttt{unk\_id=1}, \texttt{bos\_id=2}, and \texttt{eos\_id=3}, under the default \texttt{nmt\_nfkc} normalization; they differ primarily in \texttt{split\_by\_whitespace}, which is \texttt{true} for SentencePiece-BPE and \texttt{false} for SP-BPE-CrossBoundary. The cross-boundary SentencePiece baseline additionally uses \texttt{treat\_whitespace\_as\_suffix=false}, \texttt{train\_extremely\_large\_corpus=true}, and \texttt{max\_sentence\_length=4192}. SuperBPE is benchmarked with the official two-stage curriculum: Stage 1 uses regex-based pretokenization, Stage 2 resumes training with cross-boundary merges enabled, and the inherited merge count is \texttt{int(vocab\_size * 0.78125)} for the benchmarked runs. The official SuperBPE tokenizer remains byte-based after regex splitting, which differs from SupraTok's character-plus-byte-fallback initialization. We report these initialization choices explicitly because base-unit choice affects fairness in tokenizer comparisons.

\section{Evaluation Protocol Details}
\label{app:eval}

All tokenizer comparisons within a direct comparison unit---a table panel or a row of the vocabulary-sweep table---use identical target vocabulary sizes and matched training subsets. English compression is evaluated on WikiText-103 validation; multilingual compression is evaluated on FLORES-200 devtest. Downstream language models are trained with nanoGPT on FineWeb-Edu. Within each backbone configuration, SupraTok and BPE are matched by training tokens rather than by raw-text documents: each condition is trained on the same FineWeb-Edu stream until it reaches the fixed token budget (10B for 12L-768d and 30B for 24L-1024d). The more compressive tokenizer therefore sees more raw text before stopping. We report means and standard deviations over three random seeds for the downstream task experiments in Tables~\ref{tab:downstream} and~\ref{tab:downstream-ablation}. The generation-quality results in Tables~\ref{tab:perplexity} and~\ref{tab:diversity} are reported for the 12L-768d evaluation run described in Section~4.8.

\section{Script-Aware Resources and Sampling}
\label{app:script}

For Chinese, Japanese, and Thai, script-aware processing applies thresholded character-bigram gating with a minimum count threshold of 50 before allowing merges.

\begin{table}[t]
\centering
\scriptsize
\resizebox{\columnwidth}{!}{%
\begin{tabular}{lrrr}
\toprule
\textbf{Language} & \textbf{Docs} & \textbf{Sentences} & \textbf{Phase-1 tokens} \\
\midrule
English & 8,000 & 276,019 & 5,680,850 \\
German & 8,000 & 280,883 & 4,719,950 \\
Dutch & 8,000 & 234,237 & 4,141,152 \\
French & 8,000 & 223,292 & 5,529,978 \\
Spanish & 8,000 & 204,417 & 5,062,536 \\
Italian & 8,000 & 188,699 & 4,772,012 \\
Russian & 8,000 & 312,747 & 4,917,576 \\
Ukrainian & 8,000 & 253,128 & 4,312,276 \\
Polish & 8,000 & 270,589 & 4,349,563 \\
Chinese & 8,000 & 241,053 & 2,700,273 \\
Japanese & 8,000 & 242,065 & 2,306,322 \\
Korean & 8,000 & 312,485 & 5,116,734 \\
Turkish & 8,000 & 211,838 & 3,516,811 \\
Finnish & 8,000 & 313,079 & 4,286,315 \\
\midrule
\textbf{Total} & \textbf{112,000} & \textbf{3,564,531} & \textbf{61,412,348} \\
\bottomrule
\end{tabular}
}
\caption{Composition of the 14-language multilingual tokenizer-training subset for the 100k-vocabulary experiments. Documents follow the multilingual benchmark loader rule: non-empty stripped text with length at least 50. Phase-1 token counts are computed with \protect\path{AdvancedPreTokenizer.pretokenize_curriculum(text, phase=1)}.}
\label{tab:appendix-e1}
\end{table}
\FloatBarrier

For Turkish and Finnish, suffix-aware weighting uses heuristic lists of 200 high-frequency suffix candidates per language, mined directly from the multilingual tokenizer-training subset and then applied through suffix matching. Arabic processing applies hamza normalization before frequency counting. For the separate 14-language multilingual benchmark in Table~\ref{tab:multilingual}, multilingual tokenizer training uses balanced CulturaX subsets with 8,000 documents per language, selected by simple heuristic sampling from the per-language training buckets without additional stratification. Appendix Table~\ref{tab:appendix-e1} reports the per-language document, sentence, and Phase-1 pretokenizer token counts for that multilingual benchmark subset. The exact suffix-list files and sampling script will accompany the code release.

\section{Parameter Accounting for Downstream Models}
\label{app:param}

For a GPT-2-style decoder with tied token embedding/output weights, learned positional embeddings of length 1,024, hidden size $d$, vocabulary size $V$, and $L$ transformer blocks, the total parameter count is
\begin{equation}
N = Vd + 1024d + L(12d^2 + 13d) + 2d.
\end{equation}
This formula matches the nanoGPT parameterization used in our downstream runs.

With $V=256{,}000$, the 12L-768d backbone contains 282,450,432 parameters and the 24L-1024d backbone contains 565,504,000 parameters. These totals explain why nominal GPT-2 124M/355M labels are misleading for the 256k-vocabulary setting.

\section{Code and Data Availability}

Source code for SupraTok training and evaluation will be released under an open-source license upon publication. Training corpora are derived from publicly available sources (Nemotron-CC for English tokenizers, FineWeb-Edu for LM pre-training, CulturaX for multilingual tokenizers, FLORES-200 for multilingual evaluation).

\end{document}